\documentclass{article}
\usepackage{spconf,amsmath,graphicx,booktabs,multirow,amsfonts,amssymb}
\usepackage{hyperref}

\usepackage{enumitem}
\setlist{nosep, leftmargin=14pt}
\usepackage[table]{xcolor}

\title{MF$^{2}$-MVQA: A Multi-stage Feature Fusion method for Medical Visual Question Answering}
\name{Shanshan Song$^{a,c}$\qquad Jiangyun Li$^{a,b,c,\dagger}$ \qquad Jing Wang$^{a,b}$ \qquad Yuanxiu Cai$^{a,b}$\qquad Wenkai Dong$^{a,b}$}

\address{$^{a}$ School of Automation and Electrical Engineering, University of Science and Technology Beijing, \\Beijing, China  $^{b}$ Key Laboratory of Knowledge Automation for Industrial Processes, Ministry of Education,\\Beijing, China
$^{c}$ Shunde Innovation School, University of Science and Technology Beijing, \\Foshan, China  $^{\dagger}$ Correspondence: leejy@ustb.edu.cn}

\begin{document}
%
\maketitle
\begin{abstract}
There is a key problem in the medical visual question answering task that how to effectively realize the feature fusion of language and medical images with limited datasets. In order to better utilize multi-scale information of medical images, previous methods directly embed the multi-stage visual feature maps as tokens of same size respectively and fuse them with text representation. However, this will cause the confusion of visual features at different stages. To this end, we propose a simple but powerful multi-stage feature fusion method, MF$^{2}$-MVQA, which stage-wise fuses multi-level visual features with textual semantics.  MF$^{2}$-MVQA achieves the State-Of-The-Art performance on VQA-Med 2019 and VQA-RAD dataset. The results of visualization also verify that our model outperforms previous work.
\end{abstract}
\begin{keywords}
Medical visual question answering, Multimodal feature fusion, Transformer
\end{keywords}
\section{Introduction}
\label{sec:intro}
Medical visual question answering (Med-VQA) is a new exploration in visual question answering, focusing on answering relevant questions raised by specific medical images. Different from natural image datasets, professional medical prior knowledge is needed in medical dataset collection. Therefore, realizing effective Med-VQA systems in limited datasets are endowed with great importance for improving the diagnostic processing and reducing the burden of professionals, such as helping patients have basic understanding of their conditions and assisting doctors in obtaining a more precise diagnosis.

\begin{figure}[thbp]
    \centering
    \includegraphics[width=7.5cm]{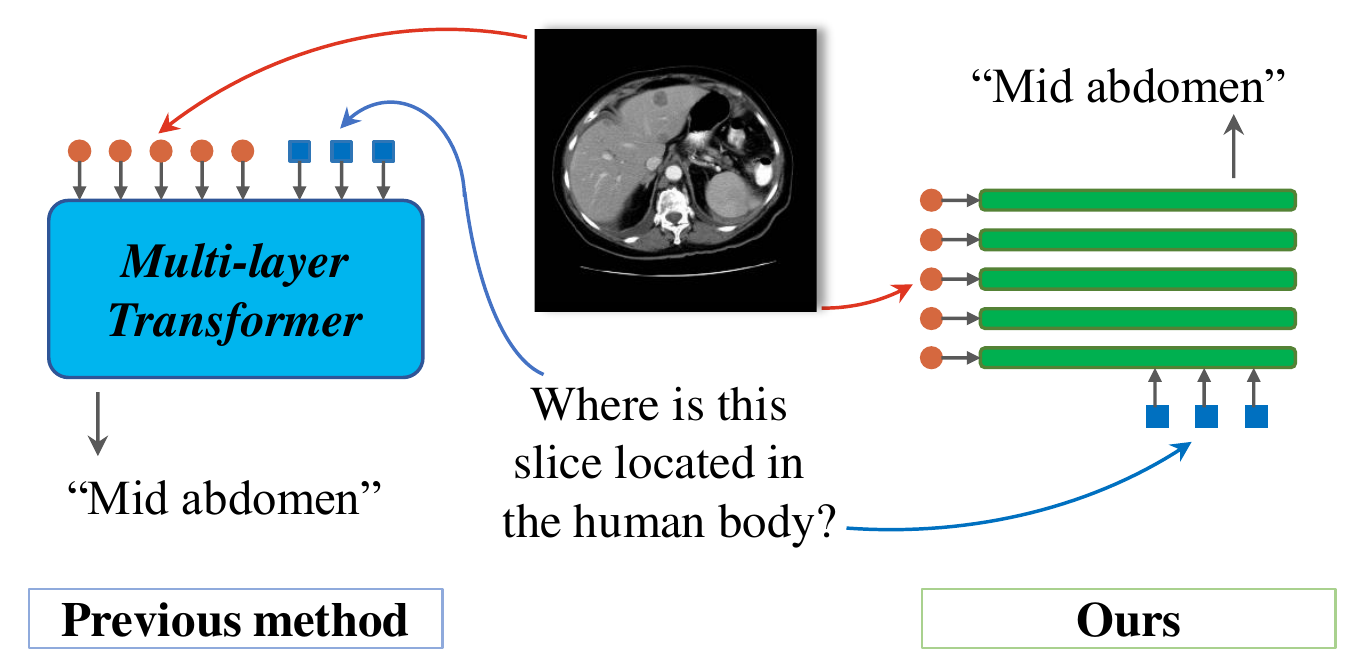}
    \caption{Structure comparison of our model and previous work for medical visual question answering task}
    \label{figure:intro}
\end{figure}
Recent works on Med-VQA mainly consist of two categories. One way is by manually designing various modules to enhance the feature's reasoning capability, including conditional reasoning framework \cite{zhan2020medical} aiming at improving the reasoning parts of Med-VQA and cross-modal self-attention module \cite{gong2021cross} to capture the long-range contextual information. Considering the strong capability of Transformer to fuse multimodal information, the other type of work makes modifications and innovations on the basis of Transformer to integrate the visual features and linguistic semantics. Ren et al. \cite{ren2020cgmvqa} proposed a multimodal Transformer architecture named CGMVQA, which is the first model utilizing different convolutional layer outputs from the pre-trained CNN-based model as the input image tokens. Khare et al. \cite{khare2021mmbert} pre-trained MMBERT model on a set of medical image-text pairs for the masked task and achieved great performance on VQA-Med 2019 \cite{abacha2019vqa} and VQA-RAD \cite{lau2018dataset} datasets. By encoding visual features from different stages to image tokens, such methods take advantage of the transformer and realize multi-scale fusion to a certain extent.

However, due to visual features of different stages containing diverse and specific information respectively, encoding them into the same resolution and then sending them together may lead to the loss of informative data. As shown in Fig.\ref{figure:intro}, based on the aforementioned problems, our work focuses on improving the fusion parts for better integration of images and language information in Transformer architecture. We propose a new Transformer-based framework for Med-VQA named MF$^{2}$-MVQA. By designing a specific masked method and layer-wise addition of image tokens for fusing multi-stage image features, our new method is capable of fully fusing multi-stage image semantics and questions. We conduct experiments on VQA-Med 2019 and VQA-RAD datasets, the results demonstrate the effectiveness of our method and the superiority over other previous methods.
\begin{figure*}[htbp]
    \centering
    \includegraphics[width=17cm]{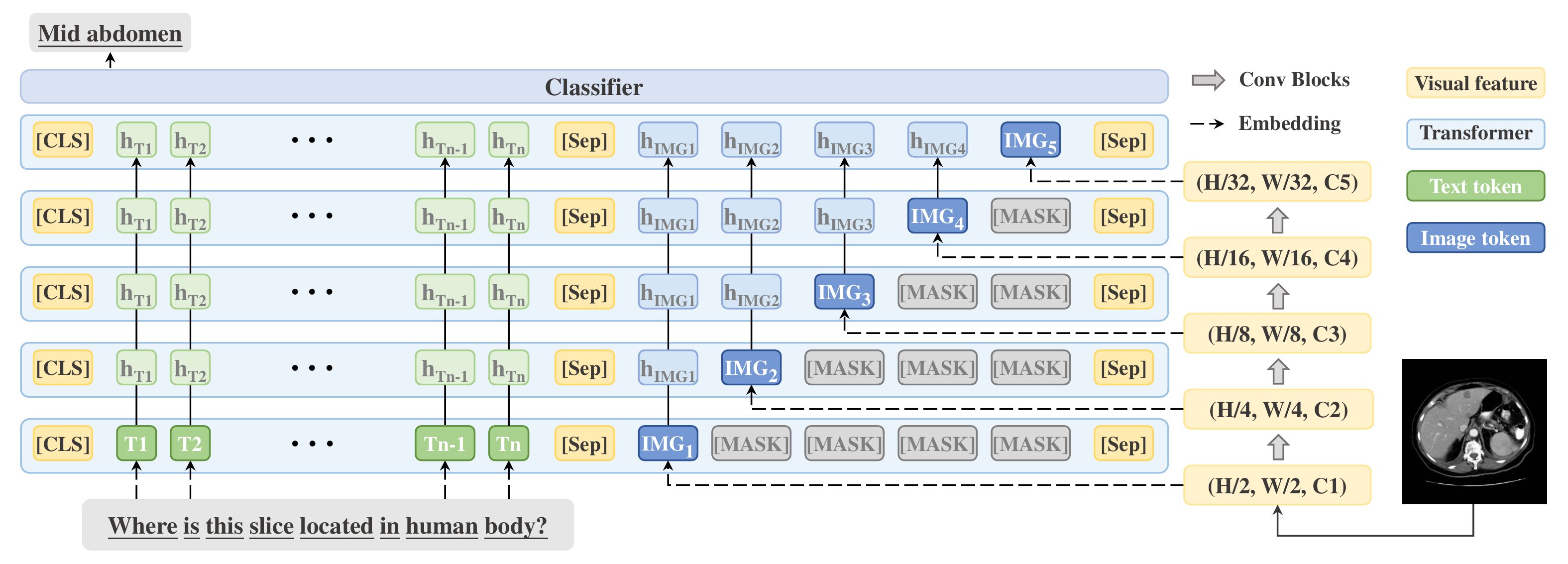}
    \caption{Overall architecture of our proposed MF$^{2}$-MVQA.The medical image is fed into the CNN encoder to obtain feature maps of different stages, and then those are added to each layer of transformer stage by stage. The visual token spaces without image feature input will be set to [MASK] tokens. The question is embedded firstly, and then the textual tokens are directly fed to the transformer structure. Finally, the classifier outputs the predicted result from candidate answers.}
    \label{figure:archi}
\end{figure*} 
\section{Methods}
\label{sec:meth}
\subsection{Transformer}
\label{ssec:trans}
MF$^{2}$-MVQA utilizes BERT-like layers as the basis of cross-modal fusion module. BERT \cite{devlin2018bert} is a multi-layer bidirectional Transformer encoder capable of modeling the dependencies of all input tokens.

For vanilla transformer structure\cite{vaswani2017attention}, it is a popular sequence transduction model. By stacking transformer encoders and decoders, models are conducted with strong sequence learning capability and efficient parallel computing. The self-attention mechanism in transformer is utilized for extracting interactions between tokens, hence the inter-modality features and the relationship between visual and linguistic representations could be obtained in our tasks. 

In each attention module, Query (Q), Key (K) and Value (V) are three crucial vectors for attention calculation. The output is a weighted sum of V, where the weight assigned to each value is the similarity between Q and its corresponding K. Specifically, the weights on V are computed by scaling the dot product of Q and V vectors following a softmax function. Meanwhile, transformers use the ensemble of self-attention modules to conduct multi-head attention, allowing the model to learn the interactions from various representation subspaces. 
\begin{equation}
    \label{eq1}
    Attention(Q,K,V) = softmax(\frac{QK^T}{\sqrt{d_k}})V
\end{equation}
With the aforementioned structures, transformer and its variants are given the advantages of cross-modal interactions. By sending feature representation tokens of two categories together into transformer, the models could achieve fusion of vision and text information. 
\begin{table*}
	\centering
	\begin{tabular}{ c c c c c c c c c c c c c }  
		\toprule[1.5pt] 
		\multirow{2}{1cm}{Method} & \multicolumn{2}{c}{Modality} & \multicolumn{2}{c}{Plane} & \multicolumn{2}{c}{Organ} & \multicolumn{2}{c}{Abnormality} & \multicolumn{2}{c}{Binary} & \multicolumn{2}{c}{Overall}\\ 
		  & Acc & BLEU & Acc & BLEU & Acc & BLEU & Acc & BLEU & Acc & BLEU & Acc & BLEU \\ 
		\midrule[1pt]     
		CGMVQA	& 80.5 & 85.6  & 80.8 & 81.3 & 72.8 & 76.9 & 1.7 & 1.7 & 75.0 & 75.0 & 60.0 & 61.9\\ 
		MMBERT NP	& 80.6 & 85.6  & 81.6 & 82.1 & 71.2 & 74.4 & 4.3 & 5.7 & 78.1 & 78.1 & 60.2 & 62.7\\
		\rowcolor{gray!18} \textbf{Ours NP}	& 73.6 & 80.4 & \textbf{86.4} & \textbf{86.4} & \textbf{72.8} & 76.1 & \textbf{6.1} & \textbf{7.8} & \textbf{85.9} & \textbf{85.9} & \textbf{62.8} & \textbf{65.0}\\
		\midrule[1pt]     
		MMBERT P	& 77.7 & 81.8  & 82.4 & 82.9 & 73.6 & 76.6 & 5.2 & 6.7 & 85.9 & 85.9 & 62.4 & 64.2\\
		\rowcolor{gray!18} \textbf{Ours P}	& \textbf{79.2} & \textbf{84.4} & \textbf{83.2} & \textbf{83.2} & 72.8 & 75.6 & \textbf{10.5} & \textbf{12.1} & \textbf{89.1} & \textbf{89.1} & \textbf{64.2} & \textbf{66.0}\\
		\midrule[1pt]     
		CGMVQA Ens	& 81.9 & 88.0 & 86.4 & 86.4 & 78.4 & 79.7 & 4.4 & 7.6 & 78.1 & 78.1 & 64.0 & 65.9\\ 
		MMBERT Ex	& 83.3 & 86.2  & 86.4 & 86.4 & 76.8 & 80.7 & 14.0 & 16.0 & 87.5 & 87.5 & 67.2 & 69.0\\
		\rowcolor{gray!18} \textbf{Ours Ex}	& \textbf{84.7} & \textbf{88.6} & \textbf{86.4} & \textbf{86.4} & 76.8 & 80.2 & \textbf{16.7} & \textbf{17.2} & \textbf{89.1} & \textbf{89.1} & \textbf{68.2} & \textbf{69.7}\\
		\bottomrule[1.5pt]   
	\end{tabular} 
	\caption{Results of our method and other comparison methods on VQA-Med 2019 dataset. NP, P, Ens and Ex represent non-pretrained model, pretrained model, ensemble models and exclusive models for different categories respectively.}
    \label{table:medvqa2019}
\end{table*}
\subsection{Overall Architecture}
\label{ssec:OA}
The overall architecture of our proposed MF$^{2}$-MVQA is illustrated in Fig.\ref{figure:archi}. Image-question pairs are taken as input to multi-scale feature fusion transformer layer and pass through a classification layer to produce corresponding answers. The whole network can be pre-trained by MLM task and acquire better performance in Med-VQA.

\textbf{Visual Feature Embedding.} A CNN-based image encoder, such as ResNet152 \cite{he2016deep} or EfficientNetv2 \cite{tan2021efficientnetv2} , is used to extract visual feature. We still choose the previous method to extract different convolutional layer outputs from image encoder because this method can obtain multi-scale information of medical image efficiently, then the feature map of each stage is resized into a token respectively through convolution layer and global average pooling layer. 

\textbf{Question Embedding.} We refer to the methods of Uniter \cite{chen2019uniter} and VilBERT \cite{lu2019vilbert} which tokenize each word in the question and use the BERT \cite{devlin2018bert} embedding layer to embed each token as a vector. To be more specific, the embedded sequence is represented by $\mathbf{w}$ = $\{w_1$, $w_2$, $\cdots$, $w_n\}$ $\in$ $\mathbb{R}^d$, where \emph{n} indicates the sequence length, and \emph{d} is the embedding dimension. After that the positional embedding is added to encode the position information. The final language representation of the question is $\{$$\hat{w_1}$, $\hat{w_2}$, $\cdots$, $\hat{w_n}$$\}$. For each of the representation at position \emph{i}, it is calculated by
\begin{equation}
    \label{eq2}
    \hat{W_i} = LayerNorm(w_i+p_i)
\end{equation}
where \emph{p$_{i}$} indicates the embedding vector at position \emph{i} and\\
\emph{LayerNorm} is a normalization function.

After the above process, the low-dimensional vector representation of the two modalities data are obtained respectively. Then the two parts feature tokens are integrated with the corresponding segment embedding and mask matrix to form a low-dimensional token sequence. After that two special tokens [CLS] and [SEP] are added for learning joint classiﬁcation feature and specifying token length.

\textbf{Cross-modal feature fusion.} Multi-layer transformers, whose layers have the same numbers with visual tokens, are adopted to conduct cross-modality feature fusion between medical image and question representation. We use a simple fusion method to fully learn cross-modal information, the specific details will be described in detail in Sec.\ref{ssec:MFF}. After sufficient feature fusion, the output pass through a classifier. The class is predicted by the output h[cls] and finally get the predicted answer from candidate items.
\subsection{Multi-stage Feature Fusion}
\label{ssec:MFF}
Recent methods directly sent whole tokens of image and text to the multi-layer transformer for feature fusion. However, such methods do not consider the difference in feature dimensions as well as the diverse and specific information in visual features from different stages. What’s more, integrating modalities together cannot effectively focus on specific image features at different stages.

Instead of feeding all visual tokens with text tokens to the five-layer transformer together, every visual token from each encoder stage is added gradually for inputting different visual features as shown in Fig.\ref{figure:archi}. To be more specific, we first set 5 tokens for all 5 visual features. In the first transformer layer, only one visual token which is from the lowest encoder stage and 4 mask tokens are sent to the first transformer layer together. In the second layer, we add a new visual token from the next stage and the number of mask tokens is 3. The remaining layers add visual tokens step by step in this way. 

Therefore, the lowest visual feature is through all transformer layers and the highest feature maps only go through 1 layer. With this simple method, we can better integrate the multi-scale information of the image and the high-level semantic information of the question.
\section{Experiment and discussion}
\label{sec:re&dis}
\begin{figure*}[htbp]
\label{figure:visual}
\centering
\includegraphics[width=16cm]{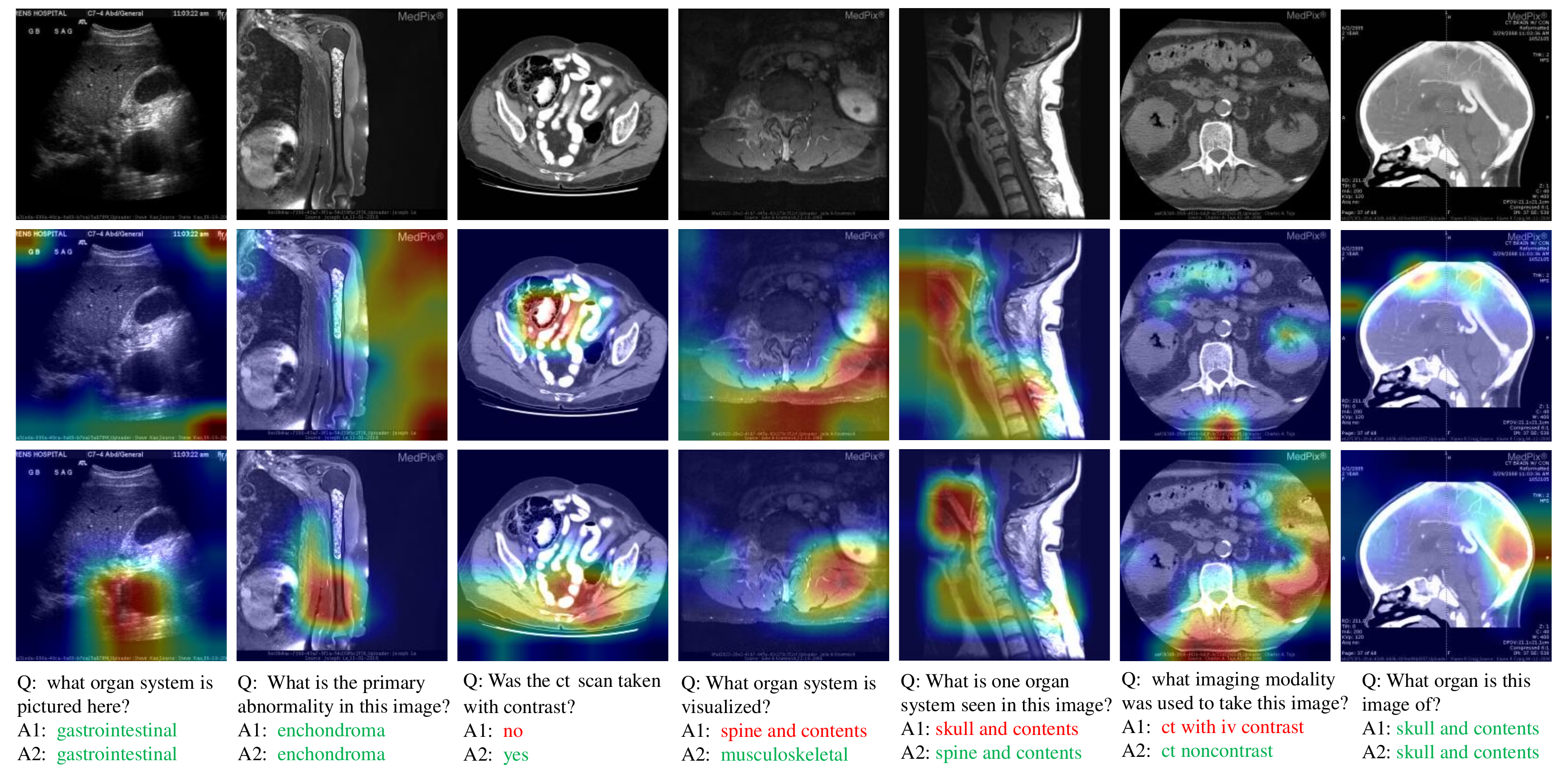}
\caption{Visualized examples of comparing MMBERT and our method on VQA-Med 2019 dataset. The first line is original images, the second line is MMBERT's heatmaps, and the third line is our visualization. Q means input question, A1 is MMBERT's prediction of answer, and A2 is our model output. Both methods are compared without loading pre-training.}
\end{figure*} 
\subsection{Dataset}
\label{ssec:data}
Radiology Objects in COntext (ROCO) dataset \cite{pelka2018radiology} is chosen to pre-train our model. The ROCO dataset consists of more than 81,000 medical image-caption pairs. We fine-tune MF$^{2}$-MVQA on two classical datasets. The first one is VQA-Med 2019. It contains 4200 medical images and 15,292 Question-Answer pairs, we split the dataset in the same way as previous work.Another dataset is VQA-RAD which contains 315 images and 3,515 corresponding questions. All questions are divided into 11 categories.
\subsection{Implementation details}
\label{ssec:imp}
The input medical images are resized to $224\times224$, and then we use EfficientNetv2 to obtain 5-stage visual feature maps. This is because even if we utilize the Efficientnetv2 block that obtains the feature map with parameter sharing, we can still reach on-par performance compared with resnet152 without parameter sharing. The feature fusion model consists of 5 transformer layers corresponding to the number of visual tokens, the hidden size is set to 768. The pretraining learning rate is 2e$^{-5}$. For fine-tuning, we use the Adam optimizer with a learning rate of 5e$^{-5}$. The rest of the parameter settings are consistent with MMBERT. All experiments were conducted on a GPU of NVIDIA 3090.

We use masked language modeling task (MLM) to pre-train our model by predicting original token of corresponding text or visual feature, which is replaced by [mask] token. For the VQA-Med 2019 dataset, we compare our methods with others under three settings respectively. First, there is no pre-training or dedicated setting, and the answer prediction is directly performed on all categories. Second, we load the pre-trained weights to predict all classes together as well. Finally, we load pre-training but train models separately for 5 different classes.
\begin{table}[h]
  \centering
  \begin{tabular}{ c c c c }
    \toprule[1.5pt]   
    Method &  Open & Closed & Overall \\
    \midrule[1pt]     
    CGMVQA &  54.2 & 78.3 & 68.7\\ 
    MMBERT P &  63.1 & 77.9 & 72.0\\    
    \textbf{Ours P} &  \textbf{63.7} & \textbf{80.1} & \textbf{73.6}\\
    \bottomrule[1.5pt]   
  \end{tabular}
  \caption{Accuracy of our method and other comparison methods on VQA-RAD dataset.}
  \label{table:vqarad} 
\end{table}
\subsection{Results and Analysis}
\label{ssec:res}
We conduct experiments on the VQA-Med 2019 dataset and compare our results with previous approaches. The results are presented in Table.\ref{table:medvqa2019}. Without data augmentation and pre-training, we considerably outperform previous methods. Compared with MMBERT, MF$^{2}$-MVQA increases the accuracy by 2.6\% and the BLEU by 2.3\%. This experiment fully demonstrates the effectiveness of our proposed multi-stage feature fusion method. Especially on the three categories of Plane, Abnormality, Yes/No, our results are even better than pre-trained MMBERT model performance.
With pretrained weights loaded, we achieve the accuracy of 64.2\% which makes a 1.8\% improvement over MMBERT and the BLEU scores of 66.0. Better results in all categories can be seen except for the organ category. Yes/No even surpassed the highest scores of other methods. In addition, we achieve state-of-the-art performance with accuracy of 68.2\% and BLEU of 69.7\% by training 5 models separately for different categories, which fully prove the effect of our Multi-stage feature fusion method.

\subsection{Qualitative Results}
\label{ssec:qua}
The heatmaps of both MMBERT and ours are visualized to gain qualitative insight of the feature fusion performance between question tokens and visual features as shown in Fig.\ref{figure:visual}. It obviously can be seen that our model (the third line) focuses on the more important regions of the medical image, which is better than MMBERT(the second line).  Even if both predicting the correct answer, our method can be relatively accurate focusing on the significant part of the medical image. This suggests that our fusion method effectively fuse semantics of multi-stage feature maps with question tokens. 

Furthermore, we also evaluate our model on the VQA-RAD dataset without setting dedicated models for different question categories, Table.\ref{table:vqarad} shows the performance comparison between our method and others. Our accuracy is up to 73.6\%, which makes a 1.6\% improvement. It is obvious that our method achieves a considerable improvement in terms of closed-ended and open-ended questions ($\uparrow$ 2.2\% and $\uparrow$ 0.6\% respectively). These results illustrate that MF$^{2}$-MVQA can get the best performance on different datasets and reveals our benefit of cross-modal inference for Med-VQA task. 
\section{Conclusion}
\label{sec:conc}
For cross-modal fusion in Med-VQA, we propose a multi-stage feature fusion model that can use transformers to better fuse medical image features and question semantics at different scales. Experiments show that our method achieves state-of-the-art on various medical question answering datasets.
\vfill
\pagebreak

\section{Acknowledgments}
\label{sec:acknowledgments}
This work was supported in part by the Natural Science Foundation of China under Grant 42201386, in part by the International Exchange Growth Program for Young Teachers of USTB under Grant QNXM20220033, and Scientific and Technological Innovation Foundation of Shunde Innovation School, USTB (BK20BE014).

\bibliographystyle{IEEEbib}
\bibliography{refs}
\end{document}